\documentclass[11pt]{article}
\usepackage{emnlp2020}
\usepackage{times}
\usepackage{url}
\usepackage{latexsym}
\usepackage{tikz}    
\usepackage{CJKutf8}
\usepackage{algorithm}
\usepackage{algpseudocode}
\usepackage[section]{placeins}
\usepackage{hyperref}
\usepackage{makecell}
\aclfinalcopy
\def\aclpaperid{41}

\title{\large Probing for Multilingual Numerical Understanding in Transformer-Based Language Models}

\author{
\begin{normalsize} 
Devin Johnson, Denise Mak, Drew Barker, Lexi Loessberg-Zahl \end{normalsize} \\
Department of Linguistics\\
University of Washington\\
{\tt \{dj1121, dpm3, barkand, lexilz\}@uw.edu}
}
\date{}
\begin{document}
\maketitle
%%%%%%%%%%%%%%%%%%%%%%%%%%

%%%%%%%%%%%%%%%%%%%%%%%%%
% ABSTRACT
%%%%%%%%%%%%%%%%%%%%%%%%%%
\begin{abstract}
Natural language numbers are an example of compositional structures, where larger numbers are composed of operations on smaller numbers. Given that compositional reasoning is a key to natural language understanding, we propose novel multilingual probing tasks tested on DistilBERT, XLM, and BERT to investigate for evidence of compositional reasoning over numerical data in various natural language number systems. By using both grammaticality judgment and value comparison classification tasks in English, Japanese, Danish, and French, we find evidence that the information encoded in these pretrained models' embeddings is sufficient for grammaticality judgments but generally not for value comparisons. We analyze possible reasons for this and discuss how our tasks could be extended in further studies.
\end{abstract}
%%%%%%%%%%%%%%%%%%%%%%%%%%

%%%%%%%%%%%%%%%%%%%%%%%%%
% INTRO
%%%%%%%%%%%%%%%%%%%%%%%%%%
\section{Introduction}
In recent years, transformer-based language models such as BERT \cite{Devlin}, XLM \cite{DBLP:journals/corr/abs-1901-07291}, and DistilBERT \cite{sanh2020distilbert} have achieved unprecedented results on a wide array of natural language understanding tasks, even when such models are trained on other tasks (i.e. transfer learning). In light of this success, there has been increased interest in investigating what particular information transformer-based language models encode in their word embeddings during pretraining that allows them to perform well in transfer learning experiments. Put into the context of this paper, we may ask: do such models gain certain linguistic/compositional understanding from pretraining? Attempts to assess such phenomena are commonly referred to as probing experiments. In this paper, we introduce a novel probing task using targeted datasets and classification tasks aimed at evaluating models’ compositional reasoning capabilities in relation to numbers\footnote{For our purposes, numbers are spelled out, i.e. written out as words such as “ninety”. } in multiple natural languages.

We choose both grammaticality judgment and value comparison tasks (Section \ref{methods}) to assess multilingual DistilBERT, XLM, and BERT\footnote{Models were chosen for their varied sizes (num. parameters) as well as our access to computing resources.} over various number systems. We argue that high performance on these tasks indicates some ability of reasoning over compositional structures, particularly over the rules generating valid compositional structures (task 1) and the resultant meanings of the structures (task 2). After probing the selected models on our tasks, we discuss explanations for the performance of models, as well as possible future extensions to this probing task schema. Additionally, studies in cognitive psychology such as \citeauthor{Miller} (\citeyear{Miller}) assert that children learning more transparent number systems (i.e. those exhibiting more regularity in their surface forms such as Japanese) have a greater counting proficiency in several tasks compared to those learning less transparent (opaque) systems, such as English or French. Although it is not the focus of our work, given the multilingual setting, we will refer to the idea of number system transparency when analyzing possible explanations of results.\footnote{Number system complexity could be the subject of its own paper. However, as a small example, we can look at ``thirty" in English and ``\begin{CJK}{UTF8}{min}
\textbf{三十}\end{CJK}" in Japanese. In English, there is no previous number such as ``three" that appears (unchanged) in the word ``thirty". In Japanese, however, the word consists of the kanji for 3 (``\begin{CJK}{UTF8}{min}
\textbf{三}\end{CJK}") and the kanji for 10 (``\begin{CJK}{UTF8}{min}
\textbf{十}\end{CJK}"). If we continue comparing in this way, we would see compositionality more clearly and regularly in Japanese's surface forms, thus forming our intuitions.}

%%%%%%%%%%%%%%%%%%%%%%%%%%

%%%%%%%%%%%%%%%%%%%%%%%%%%
% RELATED WORK
%%%%%%%%%%%%%%%%%%%%%%%%%%
\section{Related Work}
Our approach is informed by previous linguistically-motivated probing studies such at those discussed in \citeauthor{Belinkov2019AnalysisMI} (\citeyear{Belinkov2019AnalysisMI}) and \citeauthor{ettinger_2} (\citeyear{ettinger_2}). Though \citeauthor{ettinger_2} (\citeyear{ettinger_2}) discusses important findings on psycholinguistic probing experiments of BERT, we find 
\citeauthor{Ettinger} (\citeyear{Ettinger}) particularly useful for our study due to its clear explanations of linguistic probing experiment setup. In their study, Ettinger et al. present methods for constructing linguistically-targeted datasets (including example sentences, as we use) and classification tasks for probing word embeddings for semantic knowledge. As one of our tasks also seeks to probe for semantic knowledge, we were able to use this setup as a rough guideline. In addition, the authors are careful to create linguistic data with sufficient diversity as not to give potentially helpful cues to their classifier which are not related to the knowledge they wish to probe for. We thus carefully create our task data in a similar manner by limiting the distribution of our data (described more later) and forbidding duplicates. 

To our knowledge, there have been few studies conducted on investigating numerical understanding specifically in transformer-based language models with a multilingual approach. However, one particularly relevant study on English comes from \citeauthor{Wallace} (\citeyear{Wallace}). Wallace et al. probe the embeddings of various models (BERT, ELMo, word2vec, etc.) using three tasks: find the maximum of a list, decode a number word to its numerical form, and add number words to produce a numerical form. In their results, the authors note that all embeddings contain some understanding, though standard embeddings perform particularly well and character-level models perform best. Although we investigate similar phenomena as Wallace et al., our methodology includes several key differences:
\begin{itemize}
    \item Our focus is first and foremost to present a novel probing task schema/data and test it on a selection of transformer-based models - not to compare performance of differing language model architectures on previously-made tasks.
    \item We seek to draw conclusions from our task performance about model weaknesses over varied languages and suggest ways in which further probing experiments in this area can be designed in the future.
    \item We assert that the inclusion of other languages besides English is an important addition to probing experiments, as variation in language structure may help point to previously-unseen weaknesses in pretrained models.
    \item We include spelled-out numbers above 100 (up to 1000), which were not used in Wallace et al. We believe having a larger range of numbers might highlight weaknesses of models in handling multiple identical tokens in one word.
    \item We use only spelled-out number words, and do not include tasks where both Arabic numerals and spelled-out words might be used. Our reasoning for this choice is our desire to leave out the possibility of models merely learning a mapping from number words to numerals in order to perform well on tasks. In this way, we hope to make our tasks/data as restrictive as possible in order that they \textit{require} a certain compositional/linguistic understanding. 
\end{itemize}

%%%%%%%%%%%%%%%%%%%%%%%%%%

%%%%%%%%%%%%%%%%%%%%%%%%%
% METHODS
%%%%%%%%%%%%%%%%%%%%%%%%%%
\section{Methods}\label{methods}
\begin{figure}[h]
    \centering 
    \includegraphics[width=.75\linewidth]{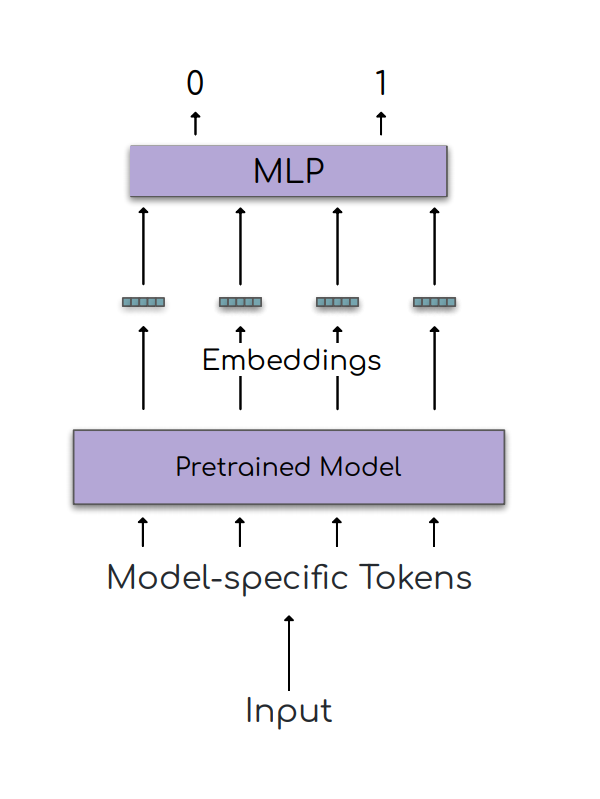}
    \caption{Probing model; Inputs can vary from sentences with numbers to standalone numbers. All weights are fixed except the connections to the MLP layer}
\label{model}
\end{figure}

We propose and perform two classification tasks in English, Danish, Japanese, and French. Task 1 is a probe for underlying syntactic information encoded in pretrained word embeddings, while task 2 is a probe of underlying semantic information. We run our tasks on all three models over two different datasets which we have generated: one where number words are inserted into sentences (e.g. ``There are seven hundred books in the library.") and one with numbers alone (e.g. ``seven-hundred"). Our probing model features a multilayer perceptron (a NN with a single hidden layer) classifier on top of the existing transformer language model architecture. In this manner, pretrained word embeddings from the language model (BERT, DistilBERT, XLM) are fed as input to the MLP classifier which itself is then trained on our tasks. A depiction of the probing model structure is shown in Figure \ref{model}. 

\subsection{Task 1: Grammaticality Judgment}
We specify the first task as follows:

\begin{itemize}
    \item Let $v \in \{bare, sentence\}$ specify the variant of our task. If $v=bare$, then only training examples with numbers \textit{not} inserted into sentences will be used for grammaticality judgments. Otherwise, only training examples with number inserted into sentences are used. A mixture of two input data types is never used.
    \item Let the training set of task 1, $T$, be defined by pairs $t_0...t_n$ where $t_i = (x_{t_i},y_{t_i})$
    \item Let $x_{t_i}$ be the input of the i'th training example $t_i$ such that $x_{t_i} = s$. $s$ is a string consisting of a number word such as ``thirty-two" or a sentence containing a number word such as ``He could eat thirty-two oranges" (depending on the value of $v$)
    \item Let $y_{t_i} \in \{0,1\}$ be the corresponding label of input $x_{t_i}$ of training sample $t_i$. $y_{t_i} = 1$ if the input string $s$ of $x_{t_i}$ is ungrammatical, otherwise $y_{t_i} = 0$.
\end{itemize}
We argue high accuracy on this task is evidence of some understanding of the underlying compositional/syntactic rules for the process which generates natural language numbers. 

\FloatBarrier

As an example, ``two hundred three and fifty” is not a number in the English language, while “five hundred” is. We also assert that the importance of using spelled-out numbers comes into play since no string of Arabic numerals is ungrammatical except a small amount of strings such as “0112". Often in written text, Arabic numerals are used in place of number words; However, given human ability to generalize compositional rules from other structures in order to learn new structures, even if it were the case that fewer number words were seen in pretraining, we would hope to see a similar compositional generalization capability present in pretrained language models' embeddings which would allow them to perform well on this task.

\subsection{Task 2: Value Comparison}
We specify the second task as follows:

\begin{itemize}
    \item Let $v \in \{bare, sentence\}$ specify the variant of our task. If $v=bare$, then only training examples with numbers \textit{not} inserted into sentences will be used for value comparsion. Otherwise, only training examples with number inserted into sentences are used. A mixture of two input data types is never used.
    \item Let the training set of task 2, $U$, be defined by pairs $u_0...u_n$ where $u_i = (x_{u_i},y_{u_i})$
    \item Let $x_{u_i}$ be the input of the i'th training example $u_i$ such that $x_{u_i} = (s_0,s_1)$. $s_0$ and $s_1$ represent bare number words or number words inserted into sentences (depending on the value of $v$).
    \item Let $y_{u_i} \in \{0,1\}$ be the corresponding label of input $x_{u_i}$ of training sample $u_i$. $y_{u_i} = 0$ if for $s_0$ and $s_1$ of $x_{u_i}$, $s_0$ refers to a value larger than that of $s_1$. $y_{u_i} = 1$ if for $s_0$ and $s_1$ of $x_{u_i}$, $s_0$ refers to a value smaller than that of $s_1$.
\end{itemize}

With this task, we take high accuracy as evidence of some understanding of the compositional semantic information carried by the number, i.e. its magnitude. For example, given the pair ($s_0=$``twelve", $s_1=$``fifteen") the correct output should be 1, since the first number in the pair is less than the second. This task is similar in form to the list maximum task of \citeauthor{Wallace} (\citeyear{Wallace}); However, notable differences include our usage of number words in sentences, our inclusion of number words above 100, and languages other than English.

%%%%%%%%%%%%%%%%%%%%%%%%%%
\begin{table*}[!htbp]
\resizebox{1\textwidth}{!}{
\begin{tabular}{lll}
\hline
\textbf{Task ($v$)}       & \textbf{Input}                                                  & \textbf{Output} \\ \hline
Task 1 (sentence)     & ``He could eat three hundred and two oranges"                & 0 ($s$ grammatical) \\ 
\hline

Task 1 (bare) & ``seventy-four six hundred and thirty-eight" & 1 ($s$ ungrammatical) \\ \hline

Task 2 (sentence)      & ``There are seven hundred and eighty-six books in the library" &  0 ($s_0$ greater) \\
                    & ``There are thirty-eight books in the library" \\   \hline

Task 2 (bare)
                    & ``five hundred" &  1 ($s_0$ less)\\
                    & ``six hundred"               & \\ \hline
                    
\end{tabular}
}
\caption{Sample of generated English data for our tasks}
\label{tab:sample data}
\end{table*}

%%%%%%%%%%%%%%%%%%%%%%%%%
% DATA
%%%%%%%%%%%%%%%%%%%%%%%%%%

\section{Data} \label{data}
Separate datasets for each variant of each task were made, each with a 60-20-20 train-validation-test split. To generate number words to create training example inputs, the python package num2words \cite{Ogawa} was used for text conversion from numerical to standard spelled-out numbers. For task 1, it was necessary to create ungrammatical numbers in each language. These ungrammatical number words were created by randomly appending grammatical number words (or parts of them) together and controlling for length. Lastly, it was also necessary to create custom sentences to insert our number words into. Eleven sentence templates were used, translated into each of our languages and verified by native speakers. A sample of our data can be seen in Table \ref{tab:sample data}. Detailed information on dataset statistics and data generation techniques can be found in appendix \ref{appendix:b}.

%%%%%%%%%%%%%%%%%%%%%%%%%%

%%%%%%%%%%%%%%%%%%%%%%%%%%
% RESULTS
%%%%%%%%%%%%%%%%%%%%%%%%%%

\section{Results}
The following sections show probing results on both tasks.
Before probing and as a precaution, we fine-tuned our models on both tasks. As we expected, we find that both tasks are learnable to accuracies above 95\%. A further discussion of fine-tuning results is left for appendix \ref{appendix:a}.

\begin{figure*}
    \centering
    \includegraphics[width=1\linewidth]{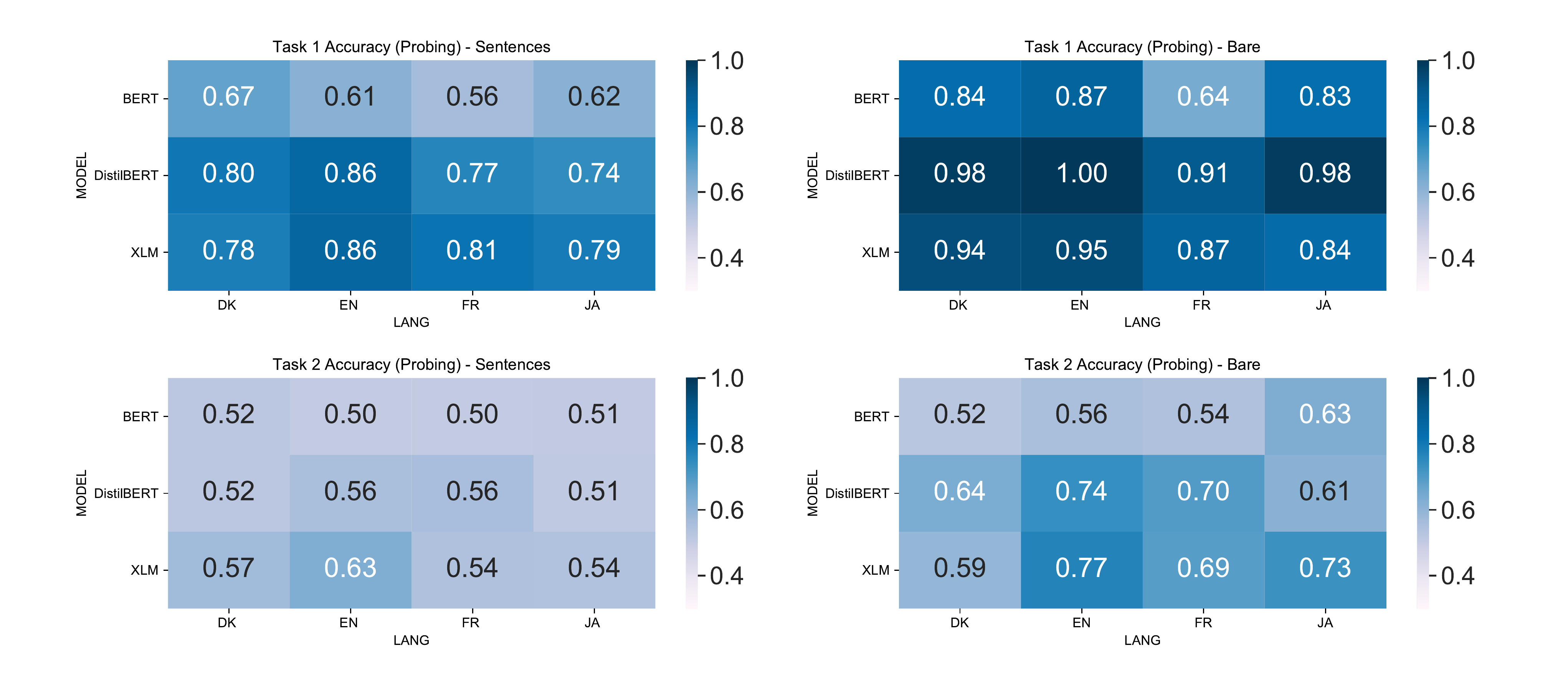}
    \caption{Task test accuracy on probing per language/model.}
    \label{fig:results}
\end{figure*}

\subsection{Task 1 Results}
Our results on task 1 (Figure \ref{fig:results}) show that the pretrained embeddings of multilingual BERT, DistilBERT and XLM seem to have sufficient information to be able to determine grammaticality of number words in Japanese, English, Danish, and French at better-than-chance performance. We thus argue that these results suggest that the pretrained embeddings of these models contain some understanding of the compositional rules for generating number words in the languages we've selected.

There are a few patterns worth noting in the results. Firstly, accuracy on bare numbers was always better than accuracy on numbers in sentences. Though accuracy on numbers in sentences was not extremely poor, our initial prediction was that they would perform better as they would resemble the type of data which the models were pretrained on (Wikipedia). Secondly, in terms of overall model performance, DistilBERT performs best (sometimes even at $100\%$ accuracy) in the majority of cases, followed by XLM and BERT. A further analysis of these patterns is left for our discussion (\ref{disc}). 

\subsection{Task 2 Results}
On our second task, we find that overall, the information in the pretrained embeddings of BERT, DistilBERT, and XLM is mostly insufficient for comparing number magnitudes in our tested languages with high accuracy. This suggests that these pretrained embeddings may struggle with understanding the compositional semantics of number words (i.e. how compositional elements in number words form to create meaning). 

As for patterns in these results, we can again see that bare number performance is always equal to or better than when numbers are inserted into sentences. Looking at bare results, DistilBERT again performs well, but this time XLM performs best on Japanese and English. A further analysis of these patterns is also left for our discussion (\ref{disc}). 

%%%%%%%%%%%%%%%%%%%%%%%%%%
% Discussion
%%%%%%%%%%%%%%%%%%%%%%%%%%

\section{Discussion} \label{disc}
\subsection{Pretraining Data}
Since this probing experiment is using pretrained transformer-based language models \cite{Wolf2019HuggingFacesTS}, we can briefly discuss pretraining methods in order to ascertain potential effects on results. Each model was pretrained using masked language modelling and all except XLM (unspecified) were pretrained on multilingual Wikipedia data. With this, one may ask whether this method of training should be expected to encode the information we probe for. Particularly, one may point to the usage of Arabic numerals in Wikipedia text and less-frequent usage of spelled-out numbers as a cause for poor performance on task 2. In imagining alternatives to Wikipedia, a consideration of Arabic numerals must still be present, which serves as a reminder that there is no \textit{perfect} pretraining set.

However, we do not believe this would prevent language models from capturing the information necessary to complete our task 2. We assert that even if these models have seen fewer spelled-out numbers, they could still learn compositional rules from other linguistic structures and generalize to our probing task. If Arabic numerals proved to be an issue, we would expect to see our results be worse across the board, not only on task 2. Thus, from our results, it seems that pretraining on Wikipedia was certainly not sufficient for encoding a highly accurate sense of number magnitude for any of the models/languages, but this was likely not due to pretraining methods.

\subsection{Worse Performance on Task 2}
Why might models have more difficulty in ascertaining magnitude of number words? For one, we believe this task is naturally more difficult than the first because of the deep semantic information necessary to succeed on it. In the first task, it may have been possible to leverage at least \textit{some} of the surface level characteristics of grammatical and ungrammatical words, whereas in the second task there is no such leveraging possible. That is to say, in the first task, a model can learn syntactic information more directly from surface level patterns. Instead, in task 2, the models need to have encoded some semantic information about the magnitude of number words, where the surface forms of these words gives less indication of their underlying meaning (except for the possibility of longer words having larger quantities, though this is not always the case). Given this is true, this may point to a weakness in models to make fine-grained semantic distinctions regarding quantities, especially when quantities are used in sentences and not left bare.

\subsection{Language Transparency}
In terms of number system transparency (as mentioned in our introduction), we loosely presumed that accuracies might follow the order of Japanese $>$ English/Danish $>$ French, with Japanese performing best given its higher transparency and French the worst due to its vigesimal number system. Again, we choose not to formally define transparency, as such a formal definition is an in-depth topic of its own. To our slight surprise, our results did not match these predictions, with rankings of performance by language varying by task and task variation. 

We argue there could be many reasons (such as pretraining data) for the unexpected results pattern. However, since there is more of a consistent performance pattern across models than there is across languages, we believe it is far more likely that differences in performance are not necessarily due to language transparency, but rather model architecture and therefore that the structural differences between these languages is not a significant contributing factor to performance patterns. If this were true (which we think is probable), this is a good sign for these models, since language structure differences are not proving to be a challenge to performance, but rather some other factor.

\subsection{Model Architecture and Performance}
One clear pattern we can see is that multilingual BERT's embeddings consistently perform much worse than both XLM and DistilBERT. This is especially clear in French results, where the gap between BERT and the other two models is sometimes more than 20 points. A relatively simple explanation for this is the size of each model in terms of number of parameters. Indeed, as model size increases, performance on both tasks increases (BERT -$>$ DistilBERT -$>$ XLM). Of course, correlation is no evidence of causation; However, if this were true, it is quite consistent with other trends in recent NLP studies. In the case of this study, we can say that bigger is (almost) always better, with XLM mostly performing best, followed by DistilBERt and BERT. Though, this is somewhat undesirable on a larger scale since, ideally, we would hope that it would not require such a large model to encode the information we probe for.

\subsection{Bare Number Words Perform Better}
On both tasks, our results also show that bare numbers performed equal to or better than numbers in sentences. We propose that this is due to the sentences creating noisiness, thus creating more difficulty for a model to know exactly where it should be looking for the necessary information to complete the tasks. This is very much the case for task 2, where we believe it would be harder to know the magnitude a sentence is referring to than merely if the sentence is grammatical. We argue that, besides adding noise, this method of probing exploits a possible weakness in masked language modeling as a pretraining method. That is, given that masked language modeling's task it to predict appropriate (grammatical) words, there may be less emphasis on learning the underlying semantics of those words, thus the better performance on task 1 sentences and worse performance on task 2 sentences.

\section{Further Work}
As this work is an exploration of a new probing method for state-of-the-art language model architecture, there are surely a number of ways to extend from it. 

Though we discussed it briefly here, exploring the architectural reasons for the shortcomings of these pretrained embeddings, especially in the case of task 2 and with sentences is an important area for future work. Indeed, in a similar task from \citeauthor{Wallace} (\citeyear{Wallace}), BERT was also found to have poor performance. In the future, several more specific probes could be designed to test for understanding of magnitude in various linguistic contexts to find strengths and weaknesses of transformer-based models. A particularly interesting case would be in testing magnitude comprehension in sentences of varying structures. Our sentence templates used in this study are few, and experimenting with other varieties could prove to be insightful.

Our experiment also made use of the idea of language transparency. We also find this to be a topic for possible further work. Namely, is there a method to reliably measure transparency of languages to predict performance on numerical understanding tasks such as these? We believe this may be possible through measuring complexities of grammars which generate number words in each language. Overall, in future extensions of this study, there is room for more languages, sentence types, task renditions, and models.

%%%%%%%%%%%%%%%%%%%%%%%%%%
% CONCLUSION
%%%%%%%%%%%%%%%%%%%%%%%%%%
\section{Conclusion}
In this paper, we introduced methods for probing the multilingual compositional reasoning capabilities of transformer-based models' pretrained embeddings over natural language numbers. From our experiments, we've shown that these pretrained embeddings show some capabilities in making grammatical judgments of number words, though they are less capable of making value comparisons. In addition, we find that results generally follow a trend based upon model size. Our results are in accord with previous work such as \citeauthor{Wallace} (\citeyear{Wallace}); However, we have also highlighted further model weaknesses through our probing methods. Therefore, the opportunities for future work, especially with a multilingual focus, are plenty.

\section*{Acknowledgements}
Thank you to all anonymous reviewers for your helpful comments. Thank you to Professor Shane Steinert-Threlkeld for your guidance and throughout all stages of this paper. Thank you to Professor Luke Zettlemoyer for your advice on our paper in its earlier stages. Thank you to the several volunteer native speakers for grammaticality judgments. \\--This work was facilitated through the use of advanced computational, storage, and networking infrastructure provided by the Hyak supercomputer system and funded by the STF at the University of Washington.--

%%%%%%%%%%%%%%%%%%%%%%%%%%

%%%%%%%%%%%%%%%%%%%%%%%%%%
% BIBLIOGRAPHY
%%%%%%%%%%%%%%%%%%%%%%%%%%
\bibliographystyle{formatting/acl_natbib.bst}
\bibliography{bib/bib_file}

%%%%%%%%%%%%%%%%%%%%%%%%%%

\newpage
\appendix
%%%%%%%%%%%%%%%%%%%%%%%%%
% Fine-Tuning
%%%%%%%%%%%%%%%%%%%%%%%%%
\section{Fine-Tuning}
\begin{figure}[h]
    \centering
    \includegraphics[width=0.8\linewidth]{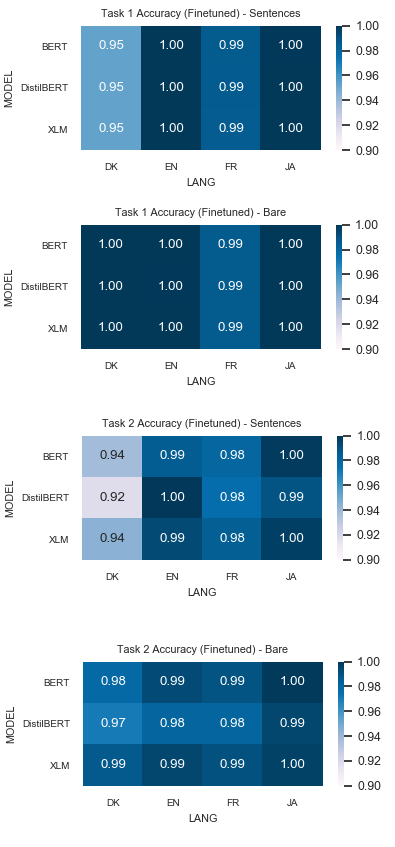}
    \caption{Fine-tuned task validation accuracy per language/model, when trained for 1 epoch.}
    \label{fig:fine_results}
\end{figure}

\begin{figure}[h]
    \centering
    \includegraphics[width=0.8\linewidth]{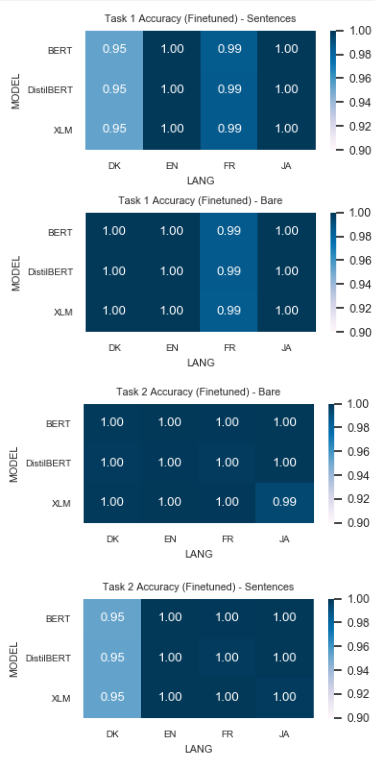}
    \caption{Fine-tuned task validation accuracy per language/model, when trained for 20 epochs.}
    \label{fig:fine_results20}
\end{figure}

The fine-tuning results shown in (Figure \ref{fig:fine_results}) are validation accuracies after one epoch of training. When trained for 20 epochs (Figure \ref{fig:fine_results20}), the models reach over 99\% accuracy on all languages and tasks except for Danish which reaches 95\% on sentence tasks. 
\label{appendix:a}
\section{Data Generation}
\label{appendix:b}

\subsection{Dataset Parameters}
The parameters below were used to generate data for each language per each model per each task variation:
\begin{itemize}
    \item Data Gen. Seed: 1
    \item Data Gen. Number Range: [0-999]
    \item Train Set Size: 30,000
    \item Validation Set Size: 10,000
    \item Test Set Size: 10,000
    \item Shuffle = True
\end{itemize}

\subsection{Task 1 Data}
For both variants of task 1 (sentences/bare), grammatical data are generated by creating random numbers then converting them to text through the num2words \cite{Ogawa} package. For the ungrammatical data, two grammatical numbers are randomly generated, both converted to text, then appended together to create an ungrammatical number. For example the ungrammatical number ``fifty-five two hundred" is the combination of ``fifty-five” and ``two hundred”. Another example made from the same original elements could be: ``two fifty-five hundred". Grammatical numbers used in splits, however, were only split such that the resulting elements were grammatical words themselves. So, for example, an non-continuous number word string like "fi-tfy te nnine" would never occur. 

Since generating numbers with appendage can naturally occur in ungrammatical number words being longer than grammatical, we control for length by limiting our set of ungrammatical number words to words that are \textit{at most} as long as the longest grammatical number. Lastly, we ensure no grammatical numbers are accidentally created in this process by keeping a list of known grammatical numbers in text form (generated by num2words) that ranges from number sufficiently higher than our generation range. For example, if our number word generation ranges from 0-1000, we would make this list of known grammatical numbers from 1-100,000,000. These number words are finally either left bare or inserted into sentences to form our $x$ inputs and are labeled 0 if grammatical and 1 if ungrammatical.

\subsection{Task 2 Data}
The data for the semantic task are generated by creating pairs of random (grammatical) number words and labeling the pair with one of two categories: 0 if the first numbers is larger than the second and 1 if the second is larger than the first. Through our process of data generation, we ensure that there are never two pairs using the same number. They are then converted to text form for input to a model. These number words are finally either left bare or inserted into sentences to form our $x$ inputs. When numbers are used in sentence templates, it is ensured that the numbers are used in the same template. For example, given the number pair ``five" and ``six", we could compare the sentences: ``There are five apples." and ``There are six apples.".

\section{Modeling}
\subsection{Pytorch Hugging Face Transformers}
We use the configurations below of transformers from Hugging Face Transformers \cite{Wolf2019HuggingFacesTS} in Pytorch on all of our reported experimental runs. Average runtimes were all around 1 hour or less.
\begin{small}
    \begin{itemize}
    \item\textbf{DistilBERT}:\begin{itemize}
        \item Class: DistilBertForSequenceClassification
        \item Config: distilbert-base-multilingual-cased
        \item Tokenizer: DistilBertTokenizer
        \item Num. Parameters: 134 million total
    \end{itemize}
    \item \textbf{BERT}\begin{itemize}
        \item Class: BertForSequenceClassification
        \item Config: base-multilingual-cased
        \item Tokenizer: BertTokenizer
        \item Num. Parameters: 110 million total
    \end{itemize}
    \item \textbf{XLM}\begin{itemize}
        \item Class: XLMForSequenceClassification
        \item Config: xlm-mlm-100-1280
        \item Tokenizer: XLMTokenizer
        \item Num. Paremeters: $\sim$550 million total (inexact)
    \end{itemize}
\end{itemize}
\end{small}

\subsection{Hyperparameters}
All experiments which produced our final results shown in the paper were run with the following hyperparemeters which were selected manually by tuning for accuracy over a validation set:
\begin{itemize}
    \item Epochs: 20 (Range: 10-20)
    \item Learning Rate: 0.00001 (Range: 1e-5 - 1e-4)
    \item Minibatch size: 32
\end{itemize}

\subsection{Infrastructure}
\begin{itemize}
    \item GPU: Nvidia Tesla P100
    \item CUDA Version: 10.1
    \item Python Version: 3.7
\end{itemize}

\subsection{Code Repository}
Our Github repository can be found
\href{https://github.com/dj1121/tlm_num_probe}{here}. Code is subject to change after publishing of this paper. Refer to the Github README for latest information.

\end{document}